\pdfoutput=1

\documentclass[11pt]{article}

\usepackage[preprint]{acl}

\usepackage{times}
\usepackage{array}
\usepackage{makecell}
\usepackage{latexsym}
\usepackage{algorithm}
\usepackage{bm}
\usepackage{algpseudocode}
\usepackage{amsmath}
\usepackage{amsfonts}  
\usepackage{amssymb}
\usepackage{multirow}
\usepackage{graphicx}
\usepackage{subcaption}
\usepackage{makecell}
\usepackage{booktabs}
\usepackage{url}
\usepackage{dsfont}
\usepackage{xspace}
\usepackage{tcolorbox}
\usepackage{pifont}
\usepackage[table]{xcolor}
\usepackage{enumitem}
\usepackage[T1]{fontenc}

\usepackage[utf8]{inputenc}

\usepackage{microtype}

\usepackage{inconsolata}

\usepackage{graphicx}

\tcbuselibrary{listings, breakable, skins, theorems}
\definecolor{front-color}{HTML}{FDEFF5}
\newtcolorbox{prompt}[1]{
    enhanced,
    drop shadow=black!5!white,
    left=4mm,
    right=4mm,
    top=1mm,
    bottom=1mm,
    boxsep=0mm,
    rounded corners,
    title=#1,
    fontupper=\normalsize\linespread{1}\fontfamily{lmr}\selectfont,
}

\title{Learning What Matters: Dynamic Dimension Selection and Aggregation for Interpretable Vision-Language Reward Modeling}

\author{
    \textbf{Qiyuan Chen}$^{1}$\thanks{\, Equal Contribution.},
    \textbf{Hongsen Huang}$^{2}$\footnotemark[1],
    \textbf{Jiahe Chen}$^{1}$,
    \textbf{Qian Shao}$^{1}$,
    \textbf{Jintai Chen}$^{3}$\\
    \textbf{Hongxia Xu}$^{1}$\footnotemark[2],
    \textbf{Renjie Hua}$^{2,4}$,
    \textbf{Chuan Ren}$^{2}$\thanks{\, Corresponding Author.},
    \textbf{Jian Wu}$^{1}$\\
    $^{1}$Zhejiang University
    $^{2}$Soochow Securities Co.,Ltd.
    $^{3}$HKUST(GZ)
    $^{4}$Nanjing University\\
    \texttt{qiyuanchen@zju.edu.cn}
}

\begin{document}
\maketitle
\begin{abstract}
Vision-language reward modeling faces a dilemma: generative approaches are interpretable but slow, while discriminative ones are efficient but act as opaque "black boxes."
To bridge this gap, we propose VL-MDR (Vision-Language Multi-Dimensional Reward), a framework that dynamically decomposes evaluation into granular, interpretable dimensions.
Instead of outputting a monolithic scalar, VL-MDR employs a visual-aware gating mechanism to identify relevant dimensions and adaptively weight them (e.g., Hallucination, Reasoning) for each specific input.
To support this, we curate a dataset of 321k vision-language preference pairs annotated across 21 fine-grained dimensions.
Extensive experiments show that VL-MDR consistently outperforms existing open-source reward models on benchmarks like VL-RewardBench.
Furthermore, we show that VL-MDR-constructed preference pairs effectively enable DPO alignment to mitigate visual hallucinations and improve reliability, providing a scalable solution for VLM alignment.
\end{abstract}

\section{Introduction}

Reward Models (RMs) serve as a cornerstone for aligning Large Vision-Language Models (LVLMs) with human preferences~\citep{christiano2017deep,ouyang2022training,yang2025qwen3}. Beyond driving Reinforcement Learning from Human Feedback (RLHF)~\citep{rafailov2023direct}, RMs are increasingly pivotal for test-time scaling strategies, such as rejection sampling, enhancing reliability without model retraining~\citep{cui2023ultrafeedback,yu2024rlhf}. As LVLMs transition towards post-training with AI-augmented synthetic data~\citep{bai2025qwen2,bai2025qwen3}, robust and scalable automated evaluation becomes essential for ensuring principled guidance throughout the model lifecycle.

\begin{figure}[t]
    \centering
    \includegraphics[width=1.0\linewidth]{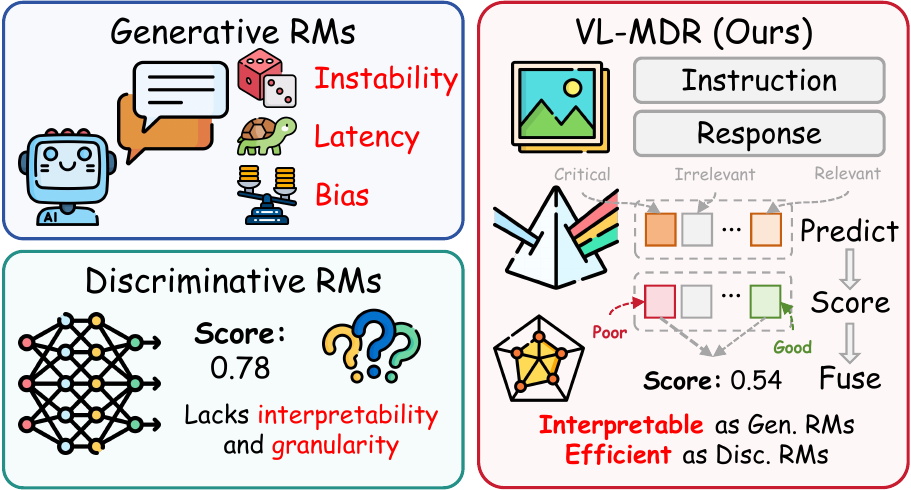}
    \caption{
    \textbf{Comparison of paradigms.} Unlike Generative RMs (high latency) and Discriminative RMs (opaque scalars), \textbf{VL-MDR} dynamically decomposes evaluation into granular dimensions, achieving both interpretability and efficiency.}
    \label{fig:compare}
    \vspace{-8mm}
\end{figure}


Existing multimodal reward modeling generally falls into two paradigms, each with distinct limitations. \textit{Generative RMs} (e.g., LLaVA-Critic~\citep{xiong2025llava}) offer interpretability via textual critiques but suffer from high latency, stochasticity, and cognitive biases (e.g., position bias)~\citep{xiong2025llava}. Conversely, \textit{Discriminative RMs} (e.g., Skywork-VL~\citep{wang2025skywork}) formulate preference learning as scalar regression. However, they operate as opaque ``black boxes'': a single scalar cannot disentangle the multidimensional nature of multimodal quality (e.g., distinguishing visual faithfulness from logical reasoning)~\citep{zang2025internlm}. Consequently, they fail to provide the granular feedback necessary for targeted model iteration.

To address this dichotomy, we take a step back to scrutinize the cognitive complexities unique to multimodal evaluation. Unlike pure text, assessing vision-language responses requires navigating a hierarchical cross-modal dependency. A response might be linguistically fluent yet visually hallucinatory; it might correctly perceive objects but fail to reason about their spatial relationships. Traditional discriminative models, by compressing these orthogonal dimensions into a single scalar, obscure the rationale behind the preference. They fail to distinguish between a model that is ``blind'' (perception failure) and one that is ``biased'' (reasoning failure), rendering the feedback opaque and difficult to steer. This motivates a question: \textit{Can we design a reward model that mirrors this granular, visually-grounded cognitive process while retaining the high throughput of discriminative scorers?}

In response, we propose the \textbf{V}ision-\textbf{L}anguage \textbf{M}ulti-\textbf{D}imensional \textbf{R}eward (\textbf{VL-MDR}) model. As shown in Figure~\ref{fig:compare}, VL-MDR reformulates evaluation as a dynamic disentanglement process. Our key insight is that a reward model should first "perceive" the intent of a multimodal query before "judging" the response.
Specifically, VL-MDR employs a three-stage mechanism:
(1) \textit{Visual-Aware Dimension Prediction}, where the model predicts relevance probabilities over dimensions and selects the Top-$k$ active dimensions for the current image-text pair;
(2) \textit{Fine-Grained Scoring}, which assesses the response quality across these targeted dimensions; and
(3) \textit{Adaptive Weighting}, which fuses these scores into a final reward based on their contextual importance.
To support this granular supervision, we curate a large-scale fine-grained preference dataset comprising approximately 321k pairs, where each pair is annotated with fine-grained preferences on the target dimensions drawn from a 21-dimension fine-grained taxonomy. This data foundation enables VL-MDR to effectively disentangle perception errors from reasoning errors, offering the interpretability of a generative judge with the efficiency of a discriminative scorer.

Extensive experiments on three multimodal reward benchmarks show that VL-MDR achieves strong and stable preference modeling with competitive performance across diverse categories. Meanwhile, VL-MDR remains efficient with a single forward pass and a lightweight multi-dimensional head, and its annotations produce higher-quality DPO preference pairs that improve downstream LVLMs and further reduce visual hallucinations.

Our main contributions are summarized as follows:
(1) We propose VL-MDR, an efficient multi-dimensional vision-language reward model that performs visual-aware dimension selection and masked adaptive aggregation to produce an interpretable reward in a single forward pass.
(2) We curate a large-scale fine-grained preference dataset of \(\sim\)321k pairs with consistent annotations over a 21-dimension hierarchical taxonomy, enabling scalable supervision for dimension-aware reward learning.
(3) Experiments on multiple benchmarks show that VL-MDR outperforms strong open-source reward models, and further provides effective preference pairs for DPO alignment to improve reliability and reduce hallucinations in downstream LVLMs.

\begin{figure*}[t]
\centering
\includegraphics[width=1.0\linewidth]{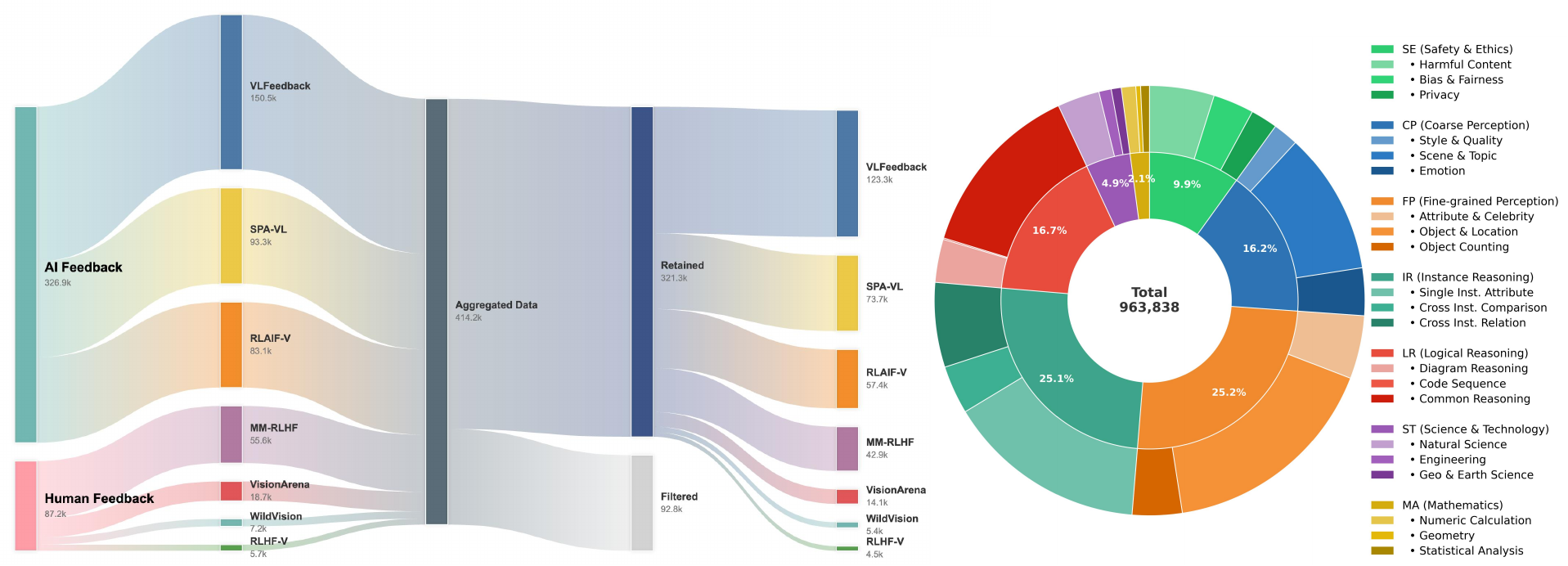}
\caption{\textbf{Data Construction Pipeline and Capability Distribution.} \textbf{Left:} We aggregate $\sim$414.2k preference samples from 7 different VLM preference datasets, grouped by supervision provenance (AI Feedback vs.\ Human Feedback), and apply our multi-model fine-grained overall-consistency filtering to retain $\sim$321.3k samples; the rightmost nodes show the retained set's source distribution. \textbf{Right:} The distribution of capability tags in the final dataset. As each sample is annotated with its top-3 relevant dimensions, the total volume of tags reaches $\sim$964k.}
\vspace{-4mm}
\label{fig:dataset_overview}

\end{figure*}

\section{Dataset Construction}
\label{sec:dataset_construction}

To facilitate fine-grained capability assessment and alignment, we construct a high-quality preference dataset containing approximately \textbf{321k} samples. As illustrated in Figure~\ref{fig:dataset_overview}, our construction pipeline integrates multi-source data integration, applies a hierarchical capability taxonomy, and enforces a multi-model fine-grained overall-consistency filtering mechanism to ensure high data quality.

\subsection{Data Integration and Taxonomy}

Our dataset is built upon a diverse, aggregated pool of approximately 414.2k preference samples collected from seven widely used VLM preference datasets. To avoid ambiguity about the provenance of supervision, we explicitly categorize these sources by the origin of their preference signals. Specifically, VLFeedback~\citep{li2023silkie}, RLAIF-V~\citep{yu2025rlaif}, and SPA-VL~\citep{zhang2025spa} provide large-scale preferences generated via AI feedback. In contrast, VisionArena~\citep{chou2025visionarena}, WildVision~\citep{lu2024wildvision}, RLHF-V~\citep{yu2024rlhf}, and MM-RLHF~\citep{zhang2025mm} primarily contain human-annotated preferences or preferences that have been human-verified, offering higher-fidelity supervision.

Moving beyond generic quality scores, we design a hierarchical capability taxonomy to precisely characterize the skills required for each sample. As detailed in Table~\ref{tab:taxonomy_definitions}, this taxonomy consists of 7 Core Capabilities which branch into 21 Fine-grained Dimensions. This structured approach allows us to decouple complex multimodal tasks into specific, interpretable skill sets, providing a granular semantic index for the entire dataset.

\begin{table}[t]
\centering
\small
\caption{\textbf{Hierarchical Capability Taxonomy.} We define seven Core Capabilities, each encompassing 3 specific Fine-grained Dimensions (21 total) to capture the micro-skills required for multimodal tasks. \textbf{Tag Ratio} reports each fine-grained dimension's share of tag slots in the retained dataset, where each sample contributes its top-3 dimensions.}
\resizebox{\columnwidth}{!}{%
\begin{tabular}{l >{\raggedright\arraybackslash}p{5.0cm} >{\raggedleft\arraybackslash}p{1.3cm}}
\toprule
\textbf{Core Capability} & \textbf{Fine-grained Dimensions} & \textbf{Tag Ratio} \\
\midrule
\textbf{Safety \& Ethics (SE)} &
\makecell[l]{Harmful Content Detection\\Bias \& Fairness\\Privacy \& Personal Information} &
\makecell[r]{4.8\%\\3.1\%\\2.0\%} \\
\midrule
\textbf{Coarse Perception (CP)} &
\makecell[l]{Style \& Quality\\Scene \& Topic\\Emotion} &
\makecell[r]{1.9\%\\10.7\%\\3.6\%} \\
\midrule
\textbf{Fine-grained Perception (FP)} &
\makecell[l]{Attribute \& Celebrity Recognition\\Object Location\\Object Counting} &
\makecell[r]{4.8\%\\16.7\%\\3.7\%} \\
\midrule
\textbf{Instance Reasoning (IR)} &
\makecell[l]{Single Instance Attribute\\Cross-instance Comparison\\Cross-instance Relation} &
\makecell[r]{15.1\%\\3.6\%\\6.4\%} \\
\midrule
\textbf{Logical Reasoning (LR)} &
\makecell[l]{Diagram Reasoning\\Code \& Sequence Reasoning\\Common Reasoning} &
\makecell[r]{3.2\%\\0.1\%\\13.3\%} \\
\midrule
\textbf{Science \& Technology (ST)} &
\makecell[l]{Natural Science\\Engineering\\Geography \& Earth Science} &
\makecell[r]{3.2\%\\0.9\%\\0.8\%} \\
\midrule
\textbf{Mathematics (MA)} &
\makecell[l]{Numeric Calculation\\Geometry\\Statistical Analysis} &
\makecell[r]{1.1\%\\0.4\%\\0.7\%} \\
\bottomrule
\end{tabular}%
}
\label{tab:taxonomy_definitions}
\vspace{-8mm}
\end{table}

\subsection{Automated Annotation and Filtering}

We employ a panel of strong VLM judges: \texttt{Qwen3-VL-235B-A22B-Instruct}, \texttt{GLM-4.5V}, and \texttt{InternVL3-78B} to automatically predict fine-grained dimensions and verify preference labels via multi-model consistency.

For each sample, we first prompt each judge to annotate the image-question pair with its top-3 fine-grained dimensions (the prompt is shown in Figure~\ref{fig:prompt_dim} in Appendix~\ref{app:prompts}). We retain a dimension annotation only when the predicted top-3 dimensions are consistent across models, ensuring stable and reproducible capability tagging.

Given the agreed target dimensions, we then prompt each judge to compare the \textit{chosen} and \textit{rejected} responses along these dimensions and output an overall preference (see Figure~\ref{fig:prompt_cmp} in Appendix~\ref{app:prompts}). A sample is retained only if (i) the overall preference is \emph{consistent across the three judges}, and (ii) the consensus preference aligns with the original ground truth (i.e., \textit{chosen} is preferred). As visualized in the Sankey diagram (Figure \ref{fig:dataset_overview}, Left), this multi-model fine-grained overall-consistency check refines the initial pool of 414.2k into a robust set of 321.3k samples, corresponding to a 77.6\% retention rate.

\subsection{Capability Distribution Analysis}

Recognizing that real-world visual tasks are rarely one-dimensional, we annotate each sample with its top-3 relevant fine-grained dimensions. This results in a total of approximately 964k capability tags, offering a comprehensive view of the dataset's composition. As shown in the sunburst chart (Figure \ref{fig:dataset_overview}, Right), the distribution is led by Fine-grained Perception (25.2\%) and Instance Reasoning (25.1\%), indicating that the majority of alignment scenarios focus on visual grounding and reasoning about object attributes or relations. Furthermore, Logical Reasoning (16.7\%) and Coarse Perception (16.2\%) remain substantial, ensuring the model is trained on multi-step tasks such as diagram analysis, common-sense deduction, and holistic understanding. Notably, Safety \& Ethics accounts for 9.9\% of the tags, providing explicit signals for harmful content, fairness, and privacy. Domain-specific categories like Science \& Technology (4.9\%) and Mathematics (2.1\%) are preserved to provide crucial supervision for expert-level capabilities. This balanced yet comprehensive distribution ensures our dataset provides precise training signals across the full spectrum of multimodal capabilities.

\begin{figure}[t]
    \centering
    \includegraphics[width=1.0\linewidth]{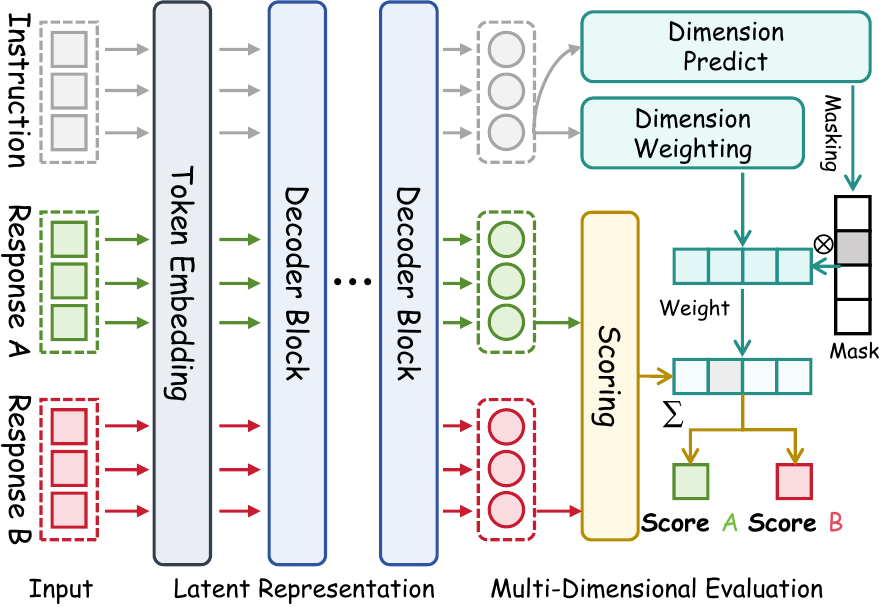}
    \caption{\textbf{Overview of the VL-MDR framework.} As shown in the diagram, the model processes the \textit{Instruction} and candidate \textit{Responses} (A and B) through a decoupled architecture. The backbone extracts distinct representations to feed three specialized heads: \textit{Dimension Predict} (identifying relevant criteria based on the instruction), \textit{Dimension Weighting} (assigning importance), and \textit{Scoring} (evaluating quality). These components are aggregated to compute the final reward signal.}
    \label{fig:vl_mdr_arch}
    \vspace{-5mm}
\end{figure}

\section{Methodology}

In this section, we present \textbf{VL-MDR} (Vision-Language Multi-Dimensional Reward), a framework designed to decompose the holistic evaluation of multimodal responses into interpretable, fine-grained components. As illustrated in Figure~\ref{fig:vl_mdr_arch}, our method decouples the evaluation criteria from response generation to simulate a human-like evaluation process: identifying \textit{what} matters via the instruction path, while assessing \textit{how} good the response is and determining \textit{how much} each aspect contributes via the response path.

\subsection{Problem Formulation}

We consider the problem of reward modeling under a pairwise preference setting. Let $x$ denote a multimodal instruction containing both visual and textual inputs. For each instruction, we are given a pair of candidate responses $(y_A, y_B)$. We define a set of $K$ evaluation dimensions, denoted as $\mathcal{D} = \{1, \dots, K\}$. The training dataset consists of tuples $(x, y_A, y_B, \mathbf{z}, \mathbf{p}, o)$, where:
$\mathbf{z} \in \{0, 1\}^K$ is the \textit{dimension relevance label}, where $z_k=1$ indicates that dimension $k$ is relevant to the instruction $x$.
$\mathbf{p} \in \{1, 0, -1\}^K$ represents the \textit{fine-grained preference} as a \emph{sparse} label: for the $k$-th dimension, $p_k=1$ implies $y_A$ is preferred over $y_B$, $p_k=-1$ implies $y_B$ is preferred, and $p_k=0$ indicates a tie; importantly, $p_k$ is defined/used only when $z_k=1$.
$o \in \{1, 0, -1\}$ represents the \textit{overall preference}, following the same semantics as $\mathbf{p}$.

\subsection{Model Architecture}

VL-MDR is built upon a pre-trained VLM backbone. A core design principle of our framework is the \textit{Query-Response Decoupling}, which posits that the evaluation criteria should be determined solely by the instruction, while the performance assessment depends on the response content.

Given the input sequence, we extract two distinct pooled representations from the backbone's last hidden states to support our decoupled evaluation design. Specifically, we obtain the \textbf{instruction representation} $\mathbf{h}^q$ from the last token of the query to capture the user's intent independent of the generated content. Simultaneously, we extract the \textbf{response representation} $\mathbf{h}^r$ from the last token of the generated response.
It is worth noting that in standard decoder-only transformers, the causal self-attention mechanism ensures that each token attends to all preceding tokens. Therefore, the final token's hidden state naturally aggregates the comprehensive information of the entire sequence, effectively encoding the full context of both the instruction and the response without the need for redundant concatenation. Based on these features, we introduce three parallel projection heads:

\paragraph{Dimension Prediction.} 
To determine the active evaluation criteria, a dimension prediction head $\phi$ maps the instruction representation to relevance logits $\mathbf{l} \in \mathbb{R}^K$:
\begin{equation}
    \mathbf{l} = \phi(\mathbf{h}^q).
\end{equation}
During inference, we first obtain relevance probabilities $\sigma(\mathbf{l})$ and then generate a \emph{Top-$k$} binary mask $\mathbf{m} \in \{0,1\}^K$ that keeps only the $k$ dimensions with the largest probabilities.

\paragraph{Fine-Grained Scoring.} 
To assess the quality of the response across all potential dimensions, a scoring head $\psi$ maps the response representation to scalar scores $\mathbf{s} \in \mathbb{R}^K$:
\begin{equation}
    \mathbf{s} = \psi(\mathbf{h}^r).
\end{equation}

\paragraph{Adaptive Aggregation.} 
To synthesize a comprehensive reward, we employ a weighting head $\omega$ that predicts raw importance logits $\mathbf{u} = \omega(\mathbf{h}^q)$. We propose a masked normalization mechanism to compute the final dimension weights $\boldsymbol{\alpha} \in \mathbb{R}^K$:
\begin{equation}
    \alpha_k = \frac{m_k \cdot \exp(u_k)}{\sum_{j=1}^K m_j \cdot \exp(u_j)}.
\end{equation}
This ensures that irrelevant dimensions (where $m_k=0$) are strictly excluded from the aggregation. The final holistic reward $R(x, y)$ is obtained by the weighted sum of the dimension scores:
\begin{equation}
    R(x, y) = \sum_{k=1}^K \alpha_k \cdot \sigma(s_k).
\end{equation}

\subsection{Hierarchical Multi-Objective Optimization}

We train VL-MDR using a hierarchical objective that jointly optimizes criteria selection, fine-grained ranking, and overall alignment.

\paragraph{Dimension Prediction Loss.} 
Since the relevance of evaluation dimensions is intrinsic to the instruction $x$, both responses $y_A$ and $y_B$ share the same ground-truth labels $\mathbf{z}$. We optimize the dimension predictor using the binary cross-entropy loss:
\begin{equation}
    \begin{aligned}
    \mathcal{L}_{\text{dim}} &= -\frac{1}{K} \sum_{k=1}^K
    \left[ z_k \log \sigma(l_k) \right. \\
    &\left. \quad + (1 - z_k) \log \left(1 - \sigma(l_k)\right) \right].
    \end{aligned}
\end{equation}

\paragraph{Unified Pairwise Ranking Loss.} 
For preference learning, we must handle both strict preference (ranking) and equality (tie) scenarios. We formulate a unified pairwise loss function $\ell(\delta, y)$ for a score difference $\delta$ and a label $y \in \{1, 0, -1\}$:
\begin{equation}
    \ell(\delta, y) = \mathbb{I}[y \neq 0] \cdot \max(0, \xi - y \cdot \delta) + \mathbb{I}[y = 0] \cdot \delta^2,
\end{equation}
where $\mathbb{I}[\cdot]$ is the indicator function and $\xi$ is the margin. The first term enforces separation when a clear preference exists, while the second term (MSE) encourages score alignment when the responses are tied.

We apply this unified loss to both the fine-grained dimensions and the overall reward. For the fine-grained scores, let $\Delta s_k = s_{A,k} - s_{B,k}$. Since fine-grained preferences are only labeled on relevant (top-3) dimensions, the ranking loss is averaged over labeled dimensions:
\begin{equation}
    \mathcal{L}_{\text{rank}} = \frac{1}{\sum z_k} \sum_{k=1}^K z_k \cdot \ell(\Delta s_k, p_k).
\end{equation}
Similarly, for the overall reward difference $\Delta R = R(x, y_A) - R(x, y_B)$, the overall loss is defined as:
\begin{equation}
    \mathcal{L}_{\text{overall}} = \ell(\Delta R, o).
\end{equation}

\paragraph{Total Objective.} 
The final training objective is a weighted combination of the three components:
\begin{equation}
    \mathcal{L}_{\text{total}} = \lambda_{\text{dim}} \mathcal{L}_{\text{dim}} + \lambda_{\text{rank}} \mathcal{L}_{\text{rank}} + \lambda_{\text{overall}} \mathcal{L}_{\text{overall}}.
\end{equation}
This multi-task formulation ensures that VL-MDR not only aligns with human preferences globally but also provides mathematically grounded justifications through its internal dimensional predictions.

\begin{table*}[htbp]
\caption{\textbf{VL-RewardBench.} Performance comparison of our reward model (VL-MDR) with existing open-source and proprietary counterparts.}
\label{tab:vl_rewardbench}
\centering
\resizebox{0.8\textwidth}{!}{%
\begin{tabular}{@{}lcccccc@{}}
\toprule
\textbf{Models} & \textbf{\#Param} & \textbf{General} & \textbf{Hallucination} & \textbf{Reasoning} & \textbf{Overall Acc} & \textbf{Macro Acc} \\
\midrule
\multicolumn{7}{c}{\textit{Proprietary Models}} \\
\midrule
GPT-4o-mini & - & 41.70 & 34.50 & 58.20 & 41.50 & 44.80 \\
GPT-4o & - & 49.10 & 67.60 & 70.50 & 65.80 & 62.40 \\
Gemini-1.5-Flash & - & 47.80 & 59.60 & 58.40 & 57.60 & 55.30 \\
Gemini-1.5-Pro & - & 50.80 & 72.50 & 64.20 & 67.20 & 62.50 \\
Claude 3.5 Sonnet & - & 43.40 & 55.00 & 62.30 & 55.30 & 53.60 \\
Claude 3.7 Sonnet & - & 68.08 & 70.70 & 60.81 & 66.31 & 66.53 \\
\midrule
\multicolumn{7}{c}{\textit{Open-Source Models}} \\
\midrule
LLaVA-OneVision-7B & 7B & 32.20 & 20.10 & 57.10 & 29.60 & 36.50 \\
Qwen2-VL-7B & 7B & 31.60 & 19.10 & 51.10 & 28.30 & 33.90 \\
Qwen2.5-VL-7B & 7B & 34.25 & 21.76 & 54.57 & 31.92 & 36.86 \\
InternVL3-8B & 8B & 60.22 & 43.93 & 62.46 & 51.00 & 55.54 \\
Llama-3.2-11B & 11B & 33.30 & 38.40 & 56.60 & 42.90 & 42.80 \\
Qwen2-VL-72B & 72B & 38.10 & 32.80 & 58.00 & 39.50 & 43.00 \\
Qwen2.5-VL-72B & 72B & 48.07 & 46.73 & 63.41 & 51.16 & 52.73 \\
InternVL3-78B & 78B & 69.61 & 52.47 & 64.35 & 57.98 & 62.15 \\
Llama-3.2-90B & 90B & 42.60 & 57.30 & 61.70 & 56.20 & 53.90 \\
\midrule
\multicolumn{7}{c}{\textit{Generative RMs}} \\
\midrule
LLaVA-Critic & 7B & 54.60 & 38.30 & 59.10 & 41.20 & 44.00 \\
UnifiedReward & 7B & 76.24 & 58.61 & 64.98 & 62.79 & 66.61 \\
UnifiedReward-Think & 7B & 77.35 & 72.50 & 65.62 & 71.45 & 71.82 \\
\midrule
\multicolumn{7}{c}{\textit{Discriminative RMs}} \\
\midrule
Skywork-VL-Reward & 7B & 65.75 & 79.84 & 60.88 & 72.98 & 68.82 \\
IXC-2.5-Reward & 7B & 80.11 & 65.29 & 60.25 & 66.16 & 68.55 \\
MM-RLHF-Reward & 7B & 45.04 & 50.45 & 57.55 & 50.15 & 51.01 \\
\textbf{VL-MDR} (Ours) & 7B & 71.27 & 75.17 & 69.09 & \textbf{73.06} & \textbf{71.84} \\
\bottomrule
\end{tabular}%
}
\vspace{-4mm}
\end{table*}

\section{Experiment}

In this section, we conduct extensive experiments to evaluate the performance and interpretability of VL-MDR. Specifically, we aim to answer the following research questions:

\begin{itemize}[leftmargin=*, noitemsep, topsep=2pt, parsep=0pt]   
    \item \textbf{RQ1 (Effectiveness):} How does VL-MDR compare with state-of-the-art discriminative and generative reward models across diverse multimodal benchmarks?
    \item \textbf{RQ2 (Granularity):} Does the proposed fine-grained dimensional supervision offer superior preference modeling performance compared to coarse or monolithic scalar rewards?
    \item \textbf{RQ3 (Efficiency):} To what extent does VL-MDR reduce computational overhead compared to generative approaches while maintaining competitive performance?
    \item \textbf{RQ4 (Utility):} Can VL-MDR be used to construct preference pairs for DPO-based alignment, improving the reliability and safety of downstream LVLM generations?
\end{itemize}

\subsection{Implementation Details and Evaluation}

We use \texttt{Qwen2.5-VL-7B-Instruct}~\citep{bai2025qwen2} as the backbone, freezing the vision tower and projector while fine-tuning the LLM and reward heads for 2 epochs with a learning rate of $5\times10^{-7}$, cosine scheduling, and 0.1 warmup. Training is performed on 8 NVIDIA H100 GPUs with per-device batch size 2 and 4 gradient accumulation steps (global batch size 64). For VL-MDR, we set the $k=3$, ranking margin $\xi=0.3$, and reward-head dropout 0.1, with $\lambda_{dim}=\lambda_{rank}=\lambda_{overall}=1.0$. We evaluate on VL-RewardBench~\citep{ruan2025vlrmbench}, Multimodal RewardBench~\citep{yasunaga2025multimodal}, and MM-RLHF-Reward Bench~\citep{zhang2025mm}, reporting Overall/Macro Average Accuracy on VL-RewardBench, and Acc/Acc+ on MM-RLHF-Reward Bench (Acc+ requires correctly ranking all response pairs within each sample).

\subsection{RQ1: Overall Effectiveness on Multimodal Reward Benchmarks}
\label{sec:rq1}

We first assess the overall effectiveness of VL-MDR by comparing it with state-of-the-art proprietary judges, open-source LVLMs, and representative generative/discriminative reward models on three diverse benchmarks (Table~\ref{tab:vl_rewardbench} and Table~\ref{tab:merged_mm_reward}). 
Across VL-RewardBench, VL-MDR consistently ranks among the strongest open-source reward models and achieves the best overall balance across categories, indicating reliable preference identification beyond skewed task distributions. 
On Multimodal RewardBench, VL-MDR remains competitive under broad evaluation axes (e.g., general correctness, knowledge, reasoning, safety, and VQA), demonstrating robust cross-domain generalization rather than overfitting to a single capability. 
Finally, on the more challenging MM-RLHF-Reward Bench, VL-MDR shows clear advantages on strict ranking criteria, reflecting improved sensitivity to subtle preference differences and hard cases.
Overall, the results verify that VL-MDR delivers strong and stable reward modeling performance across heterogeneous multimodal settings, while remaining competitive with both generative and discriminative baselines.

\begin{table}[h]
\centering
\caption{\textbf{Ablation study on supervision granularity and gating}. \textbf{Gran.} denotes the number of supervision dimensions; $\mathcal{L}_{\text{rank}}$ indicates whether per-dimension ranking supervision is used; \textbf{Gate} indicates instruction-aware dimension gating. Results are reported on VL-RewardBench (Overall/Macro accuracy).}
\label{tab:ablation}
\resizebox{\linewidth}{!}{
\begin{tabular}{l|ccc|cc}
\toprule
\multirow{2}{*}{\textbf{Method}} & \multicolumn{3}{c|}{\textbf{Config}} & \multicolumn{2}{c}{\textbf{Results}} \\
 & \textbf{Gran.} & $\mathcal{L}_{\text{rank}}$ & \textbf{Gate} & \textbf{Overall} & \textbf{Macro} \\
\midrule
Scalar & 1 & $-$ & $-$ & 64.55 & 60.87 \\
Implicit & 21 & $-$ & \checkmark & 68.72 & 65.94 \\
\midrule
Coarse-7D & 7 & \checkmark & \checkmark & 67.12 & 64.20 \\
Fine w/o Gate & 21 & \checkmark & $-$ & 69.77 & 68.51 \\
\midrule
\textbf{VL-MDR} & \textbf{21} & \textbf{\checkmark} & \textbf{\checkmark} & \textbf{70.81} & \textbf{69.96} \\
\bottomrule
\end{tabular}
}
\vspace{-2mm}
\end{table}

\begin{table*}[htbp]
\centering
\caption{\textbf{Performance on Multimodal RewardBench (Left) and MM-RLHF-Reward Bench (Right).} Comparison of our reward model (VL-MDR) with existing open-source and proprietary counterparts.}
\label{tab:merged_mm_reward}
\resizebox{\textwidth}{!}{%
\begin{tabular}{lccccccccccccccccc}
\toprule
\multirow{3}{*}{\textbf{Model}} & \multirow{3}{*}{\textbf{\#Param}} & \multicolumn{8}{c}{\textbf{Multimodal RewardBench}} & \multicolumn{7}{c}{\textbf{MM-RLHF-Reward Bench}} \\ \cmidrule(lr){3-10} \cmidrule(lr){11-17}
 &  & \multirow{2}{*}{\textbf{Overall}} & \multicolumn{2}{c}{\textbf{General}} & \multirow{2}{*}{\textbf{Know.}} & \multicolumn{2}{c}{\textbf{Reasoning}} & \multirow{2}{*}{\textbf{Safety}} & \multirow{2}{*}{\textbf{VQA}} & \multirow{2}{*}{\textbf{Mcq}} & \multirow{2}{*}{\textbf{Long}} & \multirow{2}{*}{\textbf{Short}} & \multirow{2}{*}{\textbf{Safety}} & \multirow{2}{*}{\textbf{Video}} & \multirow{2}{*}{\textbf{Acc}} & \multirow{2}{*}{\textbf{Acc+}} \\ \cmidrule(lr){4-5}\cmidrule(lr){7-8}
 &  &  & \textbf{Corr.} & \textbf{Pref.} &  & \textbf{Math} & \textbf{Code} &  &  &  &  &  &  &  &  &  \\
\midrule
\multicolumn{17}{c}{\textit{Proprietary Models}} \\
\midrule
GPT-4o & - & 70.8 & 62.6 & 69.0 & 72.0 & 67.6 & 62.1 & 74.8 & 87.2 & 64.3 & 78.4 & 44.1 & 56.3 & 40.0 & 58.2 & 26.1 \\
Claude 3.5 Sonnet & - & 71.5 & 62.6 & 67.8 & 73.9 & 68.6 & 65.1 & 76.8 & 85.6 & 64.3 & 67.6 & 55.9 & 65.6 & 60.0 & 62.9 & 26.1 \\
Claude 3.7 Sonnet & - & 71.9 & 58.4 & 60.7 & 78.1 & 76.3 & 71.3 & 72.0 & 86.8 & 66.7 & 91.9 & 91.2 & 87.5 & 76.0 & 82.4 & 65.2 \\
\midrule
\multicolumn{17}{c}{\textit{Generative RMs}} \\
\midrule
LLaVA-Critic & 7B & 54.1 & 50.1 & 53.2 & 51.3 & 49.2 & 49.3 & 78.0 & 52.4 & 42.9 & 75.7 & 47.1 & 53.1 & 48.0 & 53.5 & 23.9 \\
UnifiedReward & 7B & 61.6 & 63.1 & 58.4 & 55.9 & 64.4 & 44.7 & 50.4 & 77.2 & 59.5 & 83.8 & 58.8 & 59.4 & 84.0 & 68.2 & 37.0 \\
UnifiedReward-Think & 7B & 66.7 & 63.9 & 68.7 & 61.1 & 65.8 & 57.7 & 55.3 & 79.5 & 73.8 & 94.6 & 79.4 & 62.5 & 84.0 & 78.8 & 54.4 \\
\midrule
\multicolumn{17}{c}{\textit{Discriminative RMs}} \\
\midrule
Skywork-VL-Reward & 7B & 65.7 & 64.5 & 62.4 & 54.3 & 70.8 & 44.7 & 62.0 & 83.5 & 50.0 & 94.6 & 70.6 & 68.8 & 72.0 & 70.6 & 58.7 \\
IXC-2.5-Reward & 7B & 66.6 & 60.7 & 64.2 & 56.8 & 63.0 & 50.5 & 89.9 & 81.1 & 52.4 & 91.9 & 67.7 & 62.5 & 88.0 & 71.2 & 50.0 \\
MM-RLHF-Reward & 7B & 67.1 & 61.7 & 67.5 & 54.3 & 58.4 & 57.9 & 92.9 & 76.8 & 83.3 & 97.3 & 73.5 & 68.8 & 88.0 & 82.3 & 63.0 \\
VL-MDR (Ours) & 7B & 69.0 & 69.8 & 68.7 & 67.8 & 78.2 & 45.2 & 51.2 & 84.4 & 83.3 & 91.9 & 79.4 & 81.3 & 92.0 & 85.3 & 69.6 \\
\bottomrule
\end{tabular}%
}
\end{table*}

\begin{figure}[h]
    \centering
    \includegraphics[width=\linewidth]{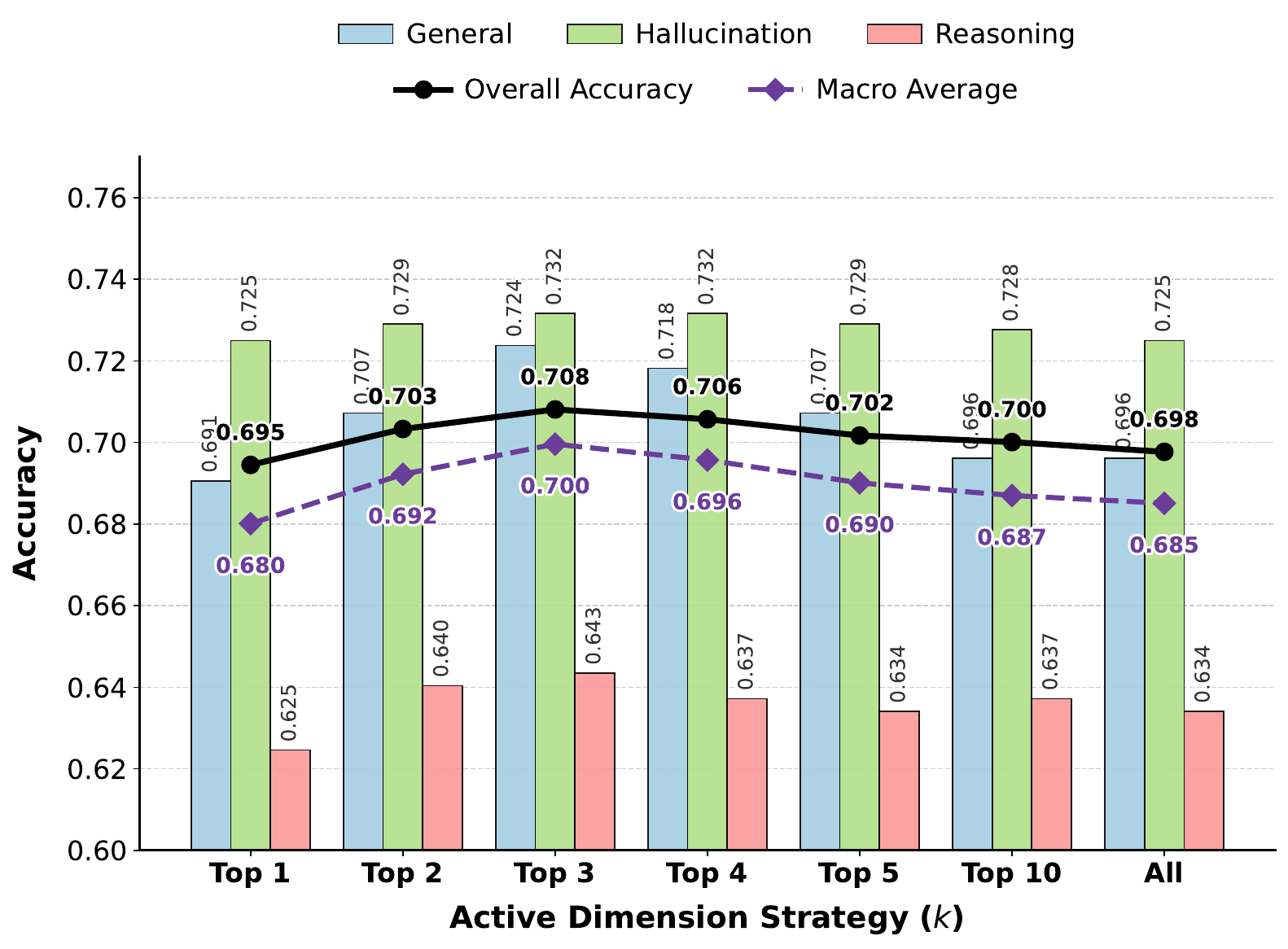}
    \caption{\textbf{Impact of Active Dimension Count ($k$).} Performance peaks at $k=3$, demonstrating that selecting a focused set of relevant dimensions strikes an optimal balance: it filters out noise from irrelevant criteria  while retaining sufficient evaluation signals.}
    \label{fig:topk_sweep}
    \vspace{-4mm}
\end{figure}

\subsection{RQ2: Impact of Granularity and Dimensional Supervision}
\label{sec:rq2}
To identify what drives VL-MDR, we conduct controlled ablations. All variants are trained on the same 200k randomly sampled subset with identical training setup, and evaluated on VL-RewardBench.

From the results in Table~\ref{tab:ablation}, we draw the following conclusions. Explicit dimensional supervision matters: removing per-dimension ranking supervision while keeping the 21-head structure (Implicit) improves over the Scalar baseline but still falls short of VL-MDR, suggesting that structural capacity alone is insufficient. Finer granularity also helps: with supervision enabled, Fine w/o Gate outperforms Coarse-7D on both Overall and Macro, indicating that 21D labels provide more specific guidance than coarse capability groups. Gating further improves performance: Fine w/o Gate underperforms VL-MDR, and the Top-$k$ sweep in Figure~\ref{fig:topk_sweep} peaks at $k=3$ before declining as more dimensions are included, showing that selecting a moderate number of relevant dimensions is optimal while aggregating too many introduces noise.

\begin{table}[t]
    \centering
    \caption{\textbf{Computational cost and performance trade-off on VL-RewardBench.} We report the total inference cost (in GPU Hours) required to evaluate the full benchmark alongside the Macro Acc. While Generative RMs suffer from high latency due to decoding, VL-MDR maintains the efficiency of Discriminative RMs.}
    \label{tab:efficiency_tradeoff}
    \resizebox{\linewidth}{!}{
    \begin{tabular}{@{}lccc@{}}
    \toprule
    \multirow{2}{*}{\textbf{Method}} & \textbf{Add. Param.} & \textbf{Cost} & \textbf{Macro Acc.} \\
    & (M) & (GPU Hours) $\downarrow$ & (\%) $\uparrow$ \\
    \midrule
    \multicolumn{4}{c}{\texttt{Generative RMs}} \\
    \midrule
    LLaVA-Critic (7B) & N/A & 0.423 $\pm$ 0.09 & 44.0 \\
    UnifiedReward (7B) & N/A & 0.399 $\pm$ 0.08 & 66.6 \\
    UnifiedReward-Think (7B) & N/A & 0.518 $\pm$ 0.11 & 71.8 \\
    \midrule
    \multicolumn{4}{c}{\texttt{Discriminative RMs}} \\
    \midrule
    Skywork-VL (7B) & 0.0036 & 0.216 $\pm$ 0.03 & 68.8 \\
    IXC-2.5-Reward (7B) & 603.98 & 0.273 $\pm$ 0.05 & 68.6 \\
    MM-RLHF-Reward (7B) & 12.85 & 0.218 $\pm$ 0.03 & 51.0 \\
    \rowcolor{gray!10} \textbf{VL-MDR (Ours)} & \textbf{17.86} & \textbf{0.218 $\pm$ 0.04} & \textbf{71.8} \\
    \bottomrule
    \end{tabular}
    }
\end{table}
\subsection{RQ3: Computational Efficiency and Parameter Analysis}
\label{sec:efficiency}

We evaluate the computational efficiency of VL-MDR on VL-RewardBench by comparing its parameter overhead and inference cost with representative reward models. For inference cost, we run each evaluation five times on a single NVIDIA H20 GPU using HuggingFace \texttt{transformers} and report the average GPU-hours in Table~\ref{tab:efficiency_tradeoff}. VL-MDR adds a lightweight multi-dimension reward head on top of \texttt{Qwen2.5-VL-7B}, including a Dimension Predictor $f_{\text{dim}}$, a Scoring Module $f_{\text{score}}$, and a Weighting Module $f_{\text{weight}}$. This introduces only \textbf{17.86M} additional parameters, i.e., \textbf{0.25\%} of the 7B backbone, leading to negligible deployment overhead compared to standard scalar discriminative reward models. For generative RMs, we report \textit{Add. Param.} as N/A since they do not introduce an explicit reward head and are obtained by fine-tuning the full backbone. In terms of inference, VL-MDR avoids autoregressive decoding and computes dimension-wise rewards in a single forward pass, whereas GenRMs typically generate critique or score tokens and thus incur higher cost as the output length grows; compared to scalar discriminative RMs, the extra computation of the MD head is marginal because it only adds a few MLP layers on top of the same backbone activations. Appendix~\ref{app:param_details} provides the detailed parameter accounting.

\subsection{RQ4: Utility for DPO Alignment}

\label{sec:rq4_dpo}

\begin{table}[t]
\centering
\scriptsize
\caption{\textbf{Image Understanding DPO Comparison.} We compare our method with generative and discriminative RMs for DPO based on LLaVA-OneVision-7B.}
\label{tab:dpo}
\resizebox{\linewidth}{!}{
\begin{tabular}{@{}lccccc@{}}
\toprule
\textbf{Method} & \textbf{LLaVABench} & \textbf{WildVision} & \textbf{LLaVABenchWilder} & \textbf{MMHal} \\
\midrule
\multicolumn{5}{c}{\texttt{Generative RMs}} \\
\midrule
OV-7B & 90.3 & 54.9 & 67.8 & 3.2 \\
w/ LLaVA-Critic & 100.3 & 67.3 & 71.6 & 3.9 \\
w/ UnifiedReward & 101.4 & 67.8 & 75.0 & 4.0 \\
w/ UnifiedReward-Think & 101.8 & 68.3 & 77.5 & 4.2 \\
\midrule
\multicolumn{5}{c}{\texttt{Discriminative RMs}} \\
\midrule
w/ Skywork-VL & 101.9 & 68.2 & 77.6 & 4.2 \\
w/ IXC-2.5-Reward & 101.6 & 67.9 & 77.0 & 4.1 \\
w/ MM-RLHF-Reward & 101.4 & 67.1 & 76.2 & 4.1 \\
\rowcolor{gray!10} \textbf{w/ VL-MDR (Ours)} & \textbf{101.9} & \textbf{68.3} & \textbf{77.9} & \textbf{4.2} \\
\bottomrule
\end{tabular}
}
\end{table}

Following the pipeline in UnifiedReward~\citep{wang2025unified}, we construct preference pairs from the LLaVA-RLHF dataset~\citep{sun2023aligning} using VL-MDR annotations. These pairs are subsequently utilized to fine-tune LLaVA-OneVision-7B (OV-7B) via DPO. All generation parameters and training configurations align strictly with~\citep{wang2025unified}, with detailed hyperparameters provided in Appendix~\ref{app:dpo_details}.
As shown in Table~\ref{tab:dpo}, VL-MDR achieves the best or comparable performance to generative and discriminative baselines across all benchmarks.
This indicates that VL-MDR's fine-grained dimensional supervision effectively filters out hallucinations and subtle errors that scalar reward models might miss, providing higher-quality signals for alignment.


\section{Related Works}

\subsection{Vision-Language Reward Modeling}

Current research on vision-language reward modeling is primarily categorized into discriminative and generative paradigms. Discriminative RMs, such as Skywork-VL Reward~\citep{wang2025skywork} and InternLM-XComposer2.5-Reward~\citep{zang2025internlm}, typically predict a single scalar reward via a linear head. Although this formulation facilitates efficient large-scale ranking, it functions as a ``black box'' by compressing diverse error types into a monolithic score, thereby obscuring the rationale behind preferences. Conversely, Generative RMs (or ``VLM-as-a-Judge''), including Prometheus-Vision~\citep{lee2024prometheus} and LLaVA-Critic~\citep{xiong2025llava}, provide natural language critiques to explain their judgments, with recent models like LLaVA-Critic R1~\citep{wang2025llava} and UnifiedReward~\citep{wang2025unified} further incorporating explicit reasoning chains for verification. However, despite offering superior interpretability, these generative methods are computationally prohibitive due to decoding latency and remain susceptible to inherent biases, such as length preference, which limits their scalability in practical alignment pipelines.

\subsection{Vision-Language Preference Data}

Recent VLM alignment has shifted from supervised fine-tuning to preference learning to improve safety and reduce hallucinations. Early works like VLFeedback~\citep{li2023silkie} and LLaVA-Critic~\citep{xiong2025llava} used proprietary models to generate ranking data for general alignment. However, RLHF-V~\citep{yu2024rlhf} demonstrated that dense, segment-level corrections are more efficient than holistic rankings for fixing fine-grained visual errors. RLAIF-V~\citep{yu2025rlaif} further improved this by using open-source models to generate high-quality feedback, reducing reliance on proprietary judges. Finally, datasets like WildVision~\citep{lu2024wildvision} and Vision Arena~\citep{chou2025visionarena} incorporate real-world user interactions to better reflect practical usage scenarios.
Nevertheless, these existing resources predominantly rely on holistic rankings or sparse textual corrections, lacking the systematic, fine-grained dimensional annotations required to explicitly disentangle orthogonal failure modes (e.g., perception vs. reasoning) for interpretable reward modeling.

\section{Conclusion}

This paper addresses the critical trade-off in vision-language reward modeling: generative models provide clear reasoning but are slow, while discriminative models are fast but act as opaque black boxes. We proposed VL-MDR, a framework that solves this by dynamically decomposing evaluation into specific, interpretable dimensions. By predicting which criteria matter for each instruction, VL-MDR achieves the clarity of a judge with the speed of a standard scorer.
To support this approach, we curated a dataset of 321k preference pairs annotated across 21 fine-grained dimensions. Experiments show that VL-MDR consistently outperforms open-source baselines on benchmarks like VL-RewardBench. Furthermore, we demonstrated that VL-MDR-constructed preference pairs effectively enable DPO alignment to improve LVLM reliability and safety. We hope this framework offers a scalable and transparent solution for future VLM alignment.

\section*{Limitations}

While our results are promising, several limitations should be noted. First, part of our supervision relies on automated multi-judge annotations and filtering; although scalable and consistent, these signals may still reflect biases or gaps of current judge models, and additional human verification could further improve reliability. Second, our 21-dimension hierarchical taxonomy offers a structured view of multimodal quality, but it is not necessarily complete and may miss criteria that matter in specific domains or contexts. Third, our dimension gating and aggregation involve practical design choices (e.g., the number of active dimensions); further study is needed to better understand robustness and calibration under distribution shifts. Finally, our experiments focus on image-text tasks and a limited set of benchmarks and backbones; future work should test broader modalities and more diverse real-world settings to better assess generalization.
\bibliography{custom}

\appendix
\newpage
\section{Dataset Analysis}
\label{sec:dataset_analysis}

This appendix presents analysis of the fine-grained judgment results, revealing patterns about dimension consistency, relationships, and difficulty.

\paragraph{Dimension Abbreviations:} Style: Style \& Quality, Scene: Scene \& Topic, Emotion: Emotion, Celebrity: Attribute \& Celebrity, Location: Object Location, Counting: Object Counting, Attribute: Single Instance Attribute, Comparison: Cross-instance Comparison, Relation: Cross-instance Relation, Diagram: Diagram Reasoning, Code: Code \& Sequence Reasoning, Common: Common Reasoning, Natural: Natural Science, Engineering: Engineering, Geo: Geo \& Earth Science, Calc: Numeric Calculation, Geometry: Geometry, Statistics: Statistical Analysis, Harmful: Harmful Content, Bias: Bias \& Fairness, Privacy: Privacy.

\subsection{Dimension Consistency}

We analyze the consistency between per-dimension judgments and the overall judgment. For each sample, we count how many of its three fine-grained dimension judgments agree with the overall preference label. Table~\ref{tab:consistency} presents the results.

\begin{table}[h]
\centering
\small
\begin{tabular}{lcc}
\toprule
Consistency Level & Count & Percentage \\
\midrule
All 3 dimensions match overall & 266,444 & 64.3\% \\
2 dimensions match overall & 108,259 & 26.1\% \\
1 dimension matches overall & 37,252 & 9.0\% \\
0 dimensions match overall & 2,177 & 0.5\% \\
\midrule
Total & 414,132 & 100\% \\
\bottomrule
\end{tabular}
\caption{Dimension-Overall Consistency}
\label{tab:consistency}
\end{table}

As shown, 64.3\% of samples have all three dimensions aligned with the overall judgment, indicating strong internal consistency. Only 0.5\% show complete disagreement. This high consistency rate validates the reliability of our fine-grained annotation approach.

\begin{figure}[h]
    \centering
    \includegraphics[width=\linewidth]{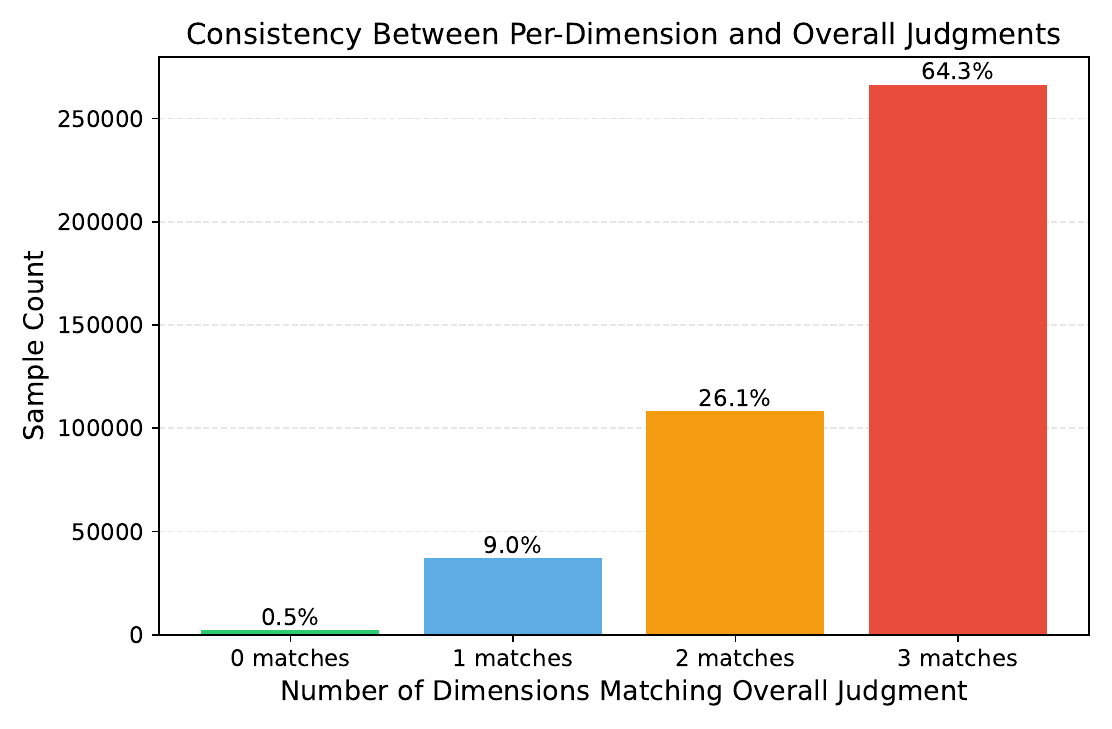}
    \caption{Dimension-overall consistency.}
    \label{fig:consistency}
\end{figure}

\subsection{Dimension Co-occurrence}

\begin{figure}[h]
    \centering
    \includegraphics[width=\linewidth]{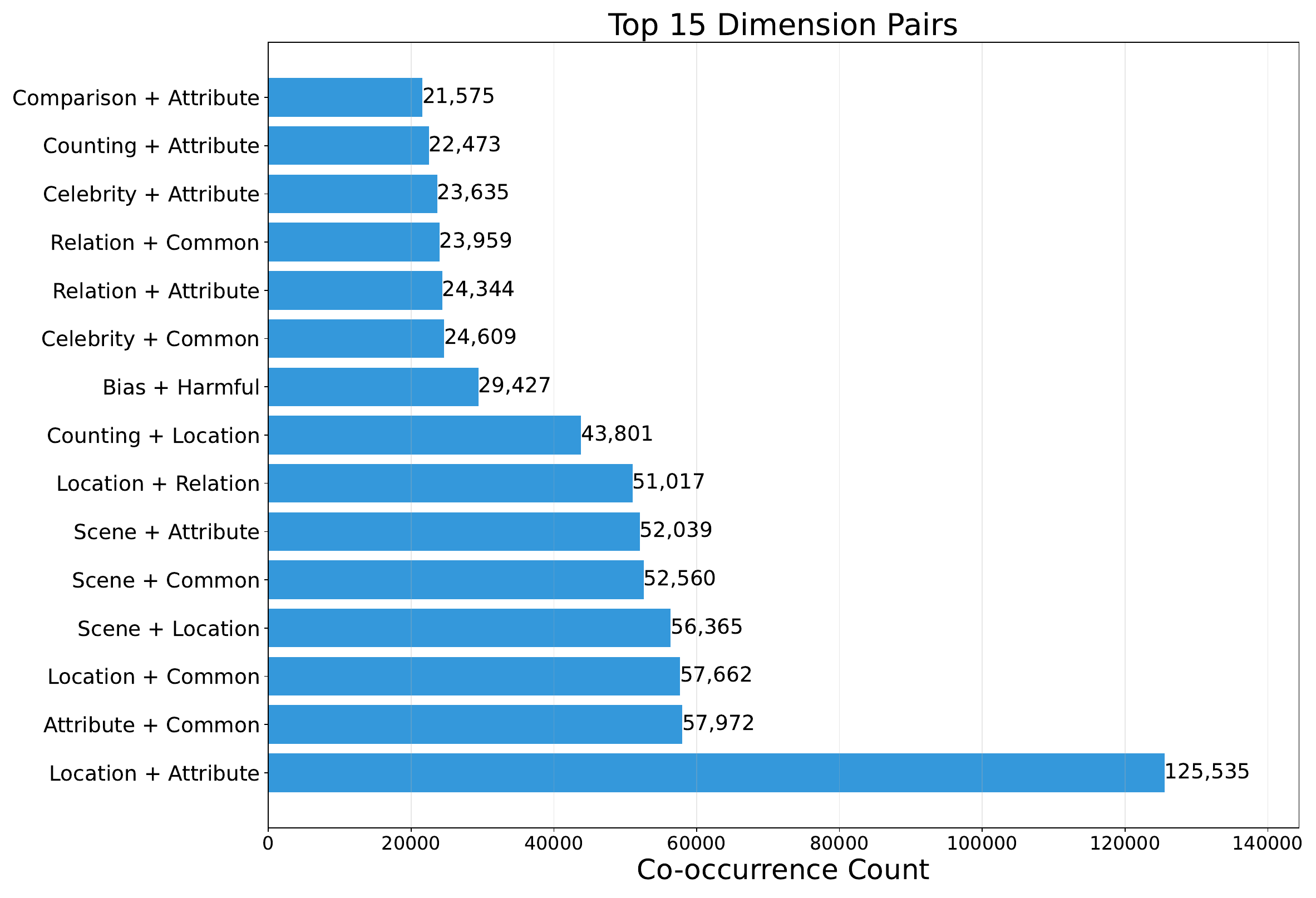}
    \caption{Top dimension co-occurrence pairs.}
    \label{fig:cooccurrence}
\end{figure}

We analyze which dimensions tend to appear together in the same samples. Figure~\ref{fig:cooccurrence} shows the top co-occurring dimension pairs.

The most frequent pair is Location + Attribute (125,535 samples), suggesting that localization and attribute description are naturally coupled tasks. Other notable pairs include Scene + Location (visual grounding) and Common + Location (reasoning about positioned objects).

\subsection{Dimension Difficulty}

Table~\ref{tab:difficulty} ranks dimensions by their tie rate, which serves as a proxy for judgment difficulty. Higher tie rates indicate more ambiguous or subjective dimensions.

\begin{table}[h]
\centering
\small
\begin{tabular}{lcc}
\toprule
Dimension & Tie Rate & Count \\
\midrule
Celebrity & 44.7\% & 61,897 \\
Calc & 19.2\% & 11,049 \\
Location & 18.3\% & 172,433 \\
Counting & 17.4\% & 44,919 \\
Attribute & 4.7\% & 182,596 \\
Geometry & 0.7\% & 5,148 \\
\bottomrule
\end{tabular}
\caption{Dimension Difficulty (Tie Rate)}
\label{tab:difficulty}
\end{table}

Celebrity recognition has the highest tie rate (44.7\%), suggesting high ambiguity in defining "celebrity" and edge cases in public figure identification. In contrast, mathematical dimensions (Geometry, Calc) have near-zero tie rates, indicating objective criteria and high inter-judge agreement.

\begin{figure}[h]
    \centering
    \includegraphics[width=\linewidth]{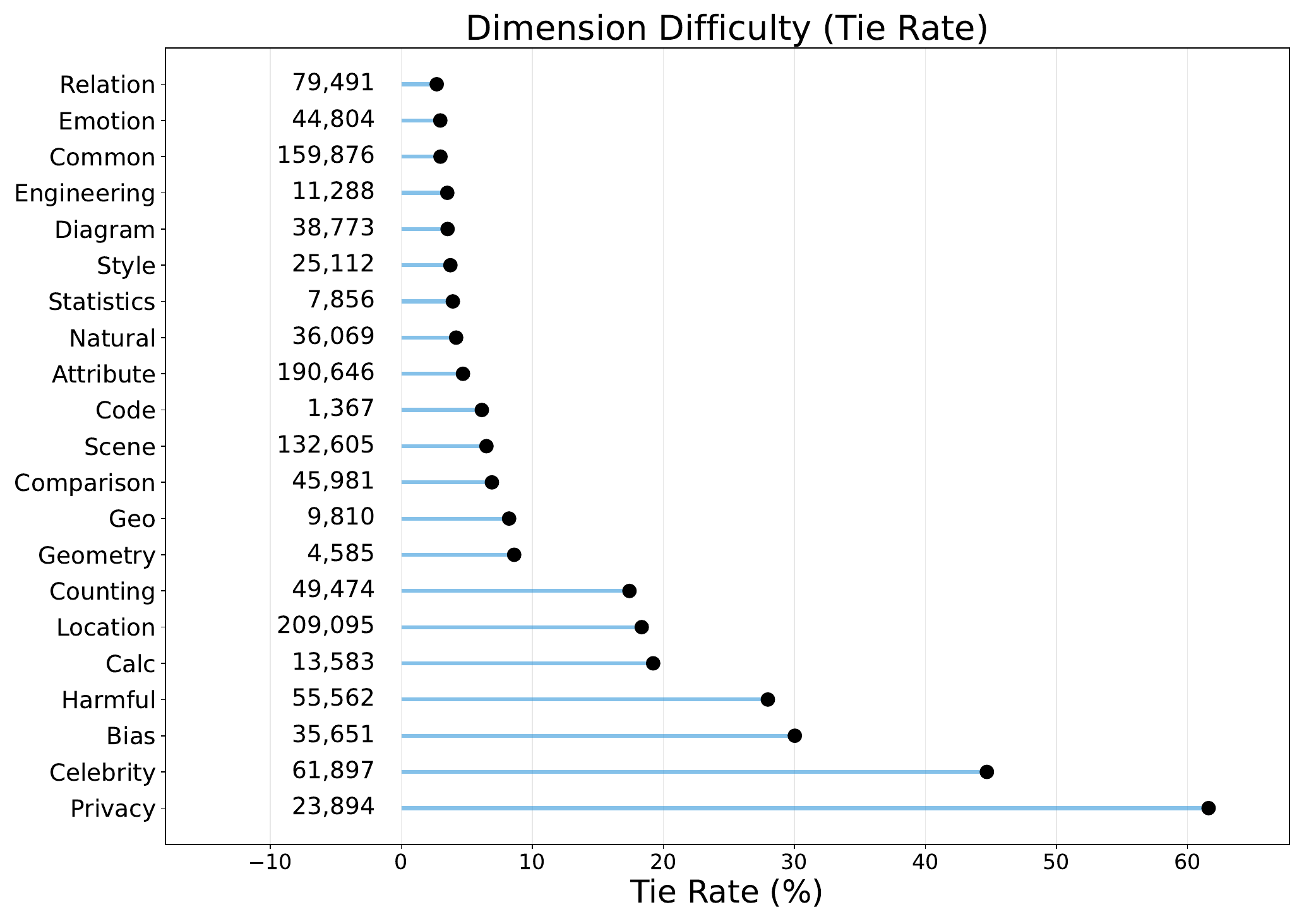}
    \caption{Dimension difficulty ranked by tie rate.}
    \label{fig:difficulty}
\end{figure}

\section{Detailed Parameter Analysis}
\label{app:param_details}

In this section, we present the detailed breakdown of the Multi-Dimensional (MD) Reward Head architecture and parameter calculations.
The backbone model is \texttt{Qwen2.5-VL-7B}, with a hidden size of $d_{in} = 3584$. The output dimension corresponds to the $K=21$ fine-grained evaluation dimensions defined in our taxonomy.
The MD Reward Head utilizes standard Linear layers (bias=False) for all transformations. The detailed layer configurations and parameter counts are listed in Table~\ref{tab:detailed_params}.

\begin{table*}[h]
    \centering
    \small
    \caption{Detailed layer-wise parameter breakdown for the VL-MDR Reward Head components.}
    \begin{tabular}{@{}lllr@{}}
    \toprule
    Component & Layer Type & Shape (Input $\to$ Output) & Params \\
    \midrule
    \multirow{4}{*}{Dimension Predictor ($f_{\text{dim}}$)}
    & Linear & $3584 \to 1024$ & 3,670,016 \\
    & Linear & $1024 \to 512$ & 524,288 \\
    & Linear & $512 \to 512$ & 262,144 \\
    & Linear & $512 \to 21$ & 10,752 \\
    \cmidrule{2-4}
    & \multicolumn{2}{l}{\textit{Subtotal}} & \textit{4.47 M} \\
    \midrule
    \multirow{5}{*}{Scoring Module ($f_{\text{score}}$)}
    & Linear & $3584 \to 2048$ & 7,340,032 \\
    & Linear & $2048 \to 1024$ & 2,097,152 \\
    & Linear & $1024 \to 1024$ & 1,048,576 \\
    & Linear & $1024 \to 512$ & 524,288 \\
    & Linear & $512 \to 21$ & 10,752 \\
    \cmidrule{2-4}
    & \multicolumn{2}{l}{\textit{Subtotal}} & \textit{11.02 M} \\
    \midrule
    \multirow{4}{*}{Weighting Module ($f_{\text{weight}}$)}
    & Linear & $3584 \to 512$ & 1,835,776 \\
    & Linear & $512 \to 512$ & 262,144 \\
    & Linear & $512 \to 512$ & 262,144 \\
    & Linear & $512 \to 21$ & 10,752 \\
    \cmidrule{2-4}
    & \multicolumn{2}{l}{\textit{Subtotal}} & \textit{2.37 M} \\
    \midrule
    Total & & & 17.86 M \\
    \bottomrule
    \end{tabular}
    \label{tab:detailed_params}
\end{table*}

\section{Details of DPO Experiment}
\label{app:dpo_details}

In this section, we provide a comprehensive description of the implementation details, training configurations, and evaluation benchmarks for the DPO experiments discussed in the main text.

\subsection{Preference Data Construction}
We adhere strictly to the data construction pipeline proposed in UnifiedReward to ensure a fair and rigorous comparison. The process utilizes image-question pairs from the LLaVA-RLHF dataset as the initial prompt source. For every prompt in the dataset, we sample six distinct candidate responses from the policy model, LLaVA-OneVision-7B~\citep{li2024llava}, using a sampling temperature of 0.7 to encourage diversity in the generated outputs.

Once the candidates are generated, we employ VL-MDR to assign a comprehensive quality score to each response. This score is derived from the weighted aggregation of our fine-grained dimensional predictions. To maximize the preference signal, we form a training pair for each prompt by designating the response with the highest reward score as the chosen sample and the response with the lowest score as the rejected sample. This filtering process results in a high-quality preference dataset comprising approximately 14,000 pairs.

\subsection{Training Configuration}
The DPO fine-tuning is conducted on the LLaVA-OneVision-7B backbone using the LLaVA-NeXT codebase. We adopt the specific hyperparameters reported in the UnifiedReward study. The model is trained for three epochs using the AdamW optimizer with $\beta_1=0.9$ and $\beta_2=0.999$. We utilize a global batch size of 128, achieved through gradient accumulation, and set the learning rate to $5 \times 10^{-7}$ with a cosine decay scheduler and a warmup ratio of 0.03. The Kullback-Leibler (KL) penalty coefficient $\beta$ for DPO is set to 0.1. All experiments are performed on a cluster of 8 NVIDIA H100 GPUs.

\subsection{Evaluation Benchmarks}

We evaluate the aligned models using the VLMEvalKit across four diverse benchmarks to assess various capabilities. LLaVABench~\citep{liu2023visual} serves as a standard metric for general visual reasoning in diverse indoor and outdoor scenes. To appraise performance in more complex and uncontrolled environments, we employ LLaVABench-Wilder~\citep{li2024llavanext-strong}. For a proxy of real-world user preference, we use WildVision, which is derived from the WildVision-Arena~\citep{lu2024wildvision} and correlates well with Chatbot Arena Elo ratings. Finally, to specifically verify the effectiveness of our model in mitigating hallucinations, we report results on MMHal-Bench~\citep{sun2023aligning}, where a higher score indicates a lower rate of hallucinatory content.

\section{Prompt Templates for Data Construction}
\label{app:prompts}

\begin{figure*}[h!]
    \centering
    \includegraphics[width=0.95\textwidth]{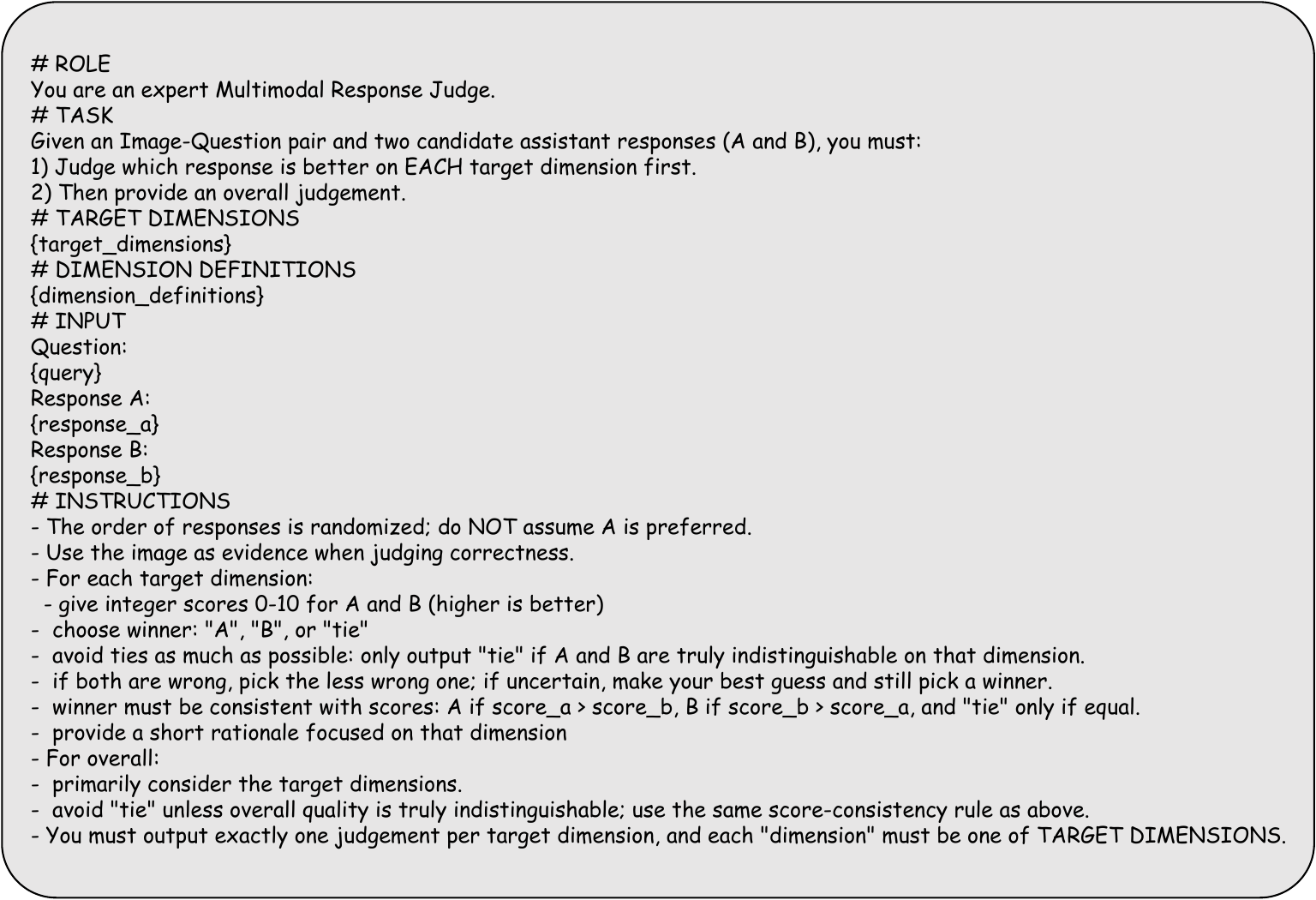}
    \caption{The prompt template used for \textbf{Fine-Grained Response Comparison}. The model evaluates two candidate responses on the specific target dimensions identified in the previous step before providing an overall preference.}
    \label{fig:prompt_cmp}
\end{figure*}

To facilitate reproducibility, we strictly follow the prompts illustrated below. Figure~\ref{fig:prompt_dim} presents the instruction used for identifying the top-3 relevant dimensions, while Figure~\ref{fig:prompt_cmp} displays the prompt used for the fine-grained comparison and overall judgment.

\begin{figure*}[h!]
    \centering
    \includegraphics[width=0.95\textwidth]{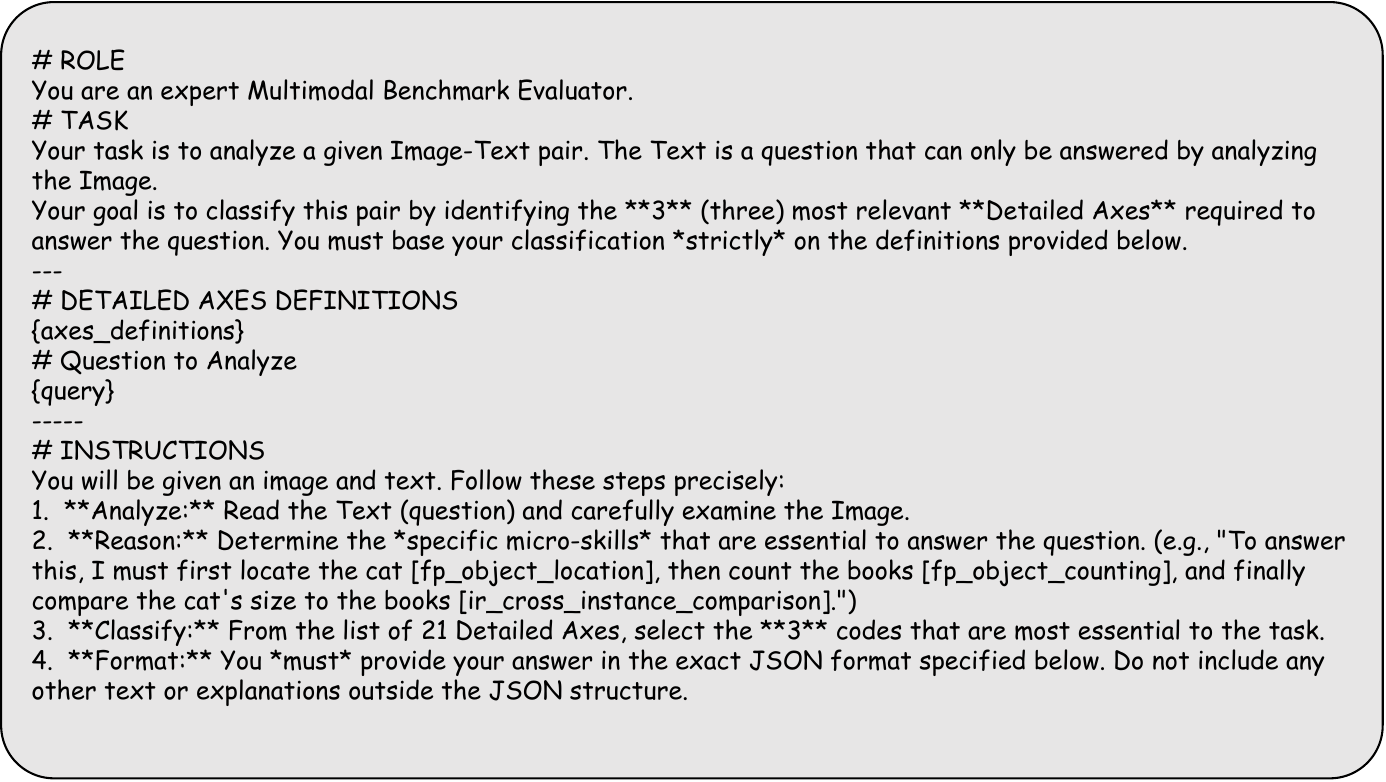} 
    \caption{The prompt template used for \textbf{Visual-Aware Dimension Prediction}. The model is instructed to analyze the image-text pair and select the top-3 relevant fine-grained axes from the defined taxonomy.}
    \label{fig:prompt_dim}
\end{figure*}

\end{document}